\documentclass{article}
\usepackage[final]{styles/colm2026_conference}
\usepackage[utf8]{inputenc}
\usepackage[T1]{fontenc}
\usepackage[protrusion=true,expansion=true,final]{microtype}
\usepackage{lineno}
\usepackage[colorlinks]{hyperref}
\definecolor{darkblue}{rgb}{0, 0, 0.5}
\hypersetup{citecolor=darkblue, linkcolor=darkblue, urlcolor=darkblue}
\usepackage[subcaption]{styles/smile}
\usepackage{styles/cme-math}
\usepackage{cancel}
\usepackage{parskip}
\usepackage{float}
\usepackage{enumitem}
\setlist{topsep=0pt,itemsep=1pt,parsep=0pt,partopsep=0pt,leftmargin=1.5em}
\usepackage{algpseudocode}

\usepackage{tikz}
\usetikzlibrary{arrows,arrows.meta,shapes,positioning,shadows,trees,decorations.pathreplacing,fit,calc}
\usepackage[most]{tcolorbox}
\usepackage{fvextra}
\DefineVerbatimEnvironment{WrapVerb}{Verbatim}{breaklines=true, breakanywhere=true}

\definecolor{darkheader}{HTML}{334155}
\definecolor{lightgray}{HTML}{FBFBFC}
\definecolor{framegray}{HTML}{E5E7EB}
\definecolor{codebg}{HTML}{F6F8FA}
\definecolor{codeframe}{HTML}{E5E7EB}
\definecolor{codetext}{HTML}{111827}
\definecolor{diffaddbg}{HTML}{E6FFEC}
\definecolor{diffaddborder}{HTML}{34D399}
\definecolor{diffaddtext}{HTML}{064E3B}
\definecolor{diffrembg}{HTML}{FFEBE9}
\definecolor{diffremborder}{HTML}{F87171}
\definecolor{diffremtext}{HTML}{7F1D1D}

\newtcolorbox{promptbox}[2][]{
    colback=lightgray, colframe=framegray, coltitle=white,
    colbacktitle=darkheader, fonttitle=\bfseries\large,
    title=#2, boxrule=0.6pt, titlerule=0pt,
    toptitle=8pt, bottomtitle=8pt, #1
}

\title{Beyond the Harness: End-to-End Optimization of Context Artifacts for Enterprise Text-to-SQL}

\author{%
Kate Gwimm\\
\texttt{kgwimm@amazon.com}
\And
Carson Eisenach\\
\texttt{ceisen@amazon.com}
}

\begin{document}

\ifcolmsubmission
\linenumbers
\fi

\maketitle

\begin{abstract}
Deploying LLMs for enterprise Text-to-SQL is bottlenecked less by the model than by what
context reaches it: business logic spans thousands of tables, and no model can ingest a
full catalog at once. We argue that the most effective place to intervene is therefore the
\emph{knowledge-base context} the model consumes, and that this context should be
\emph{constructed} from historical usage rather than tuned for as a fixed input.
Using a query-DAG decomposition--the same family of intermediates that enterprise
benchmarks like BEAVER annotate, here recovered from production SQL--we compare the value of oracle query
graphs versus retrieved knowledge-base context. In this ablation, retrieved
knowledge-base context provides the largest marginal improvement when added to the full
oracle graph.
Building on this, we optimize a distillation procedure that turns historical query
profiles into reusable SQL reference cards. On a benchmark of 5176 production queries from a major online retailer,
optimizing these context artifacts yields larger gains (${\sim}12$--$25\%$ AST
similarity) than optimizing the retrieval harness (${\sim}3$--$12\%$). On the public
BEAVER benchmark, which lacks the production-usage signals available in our internal
setting, the picture is more mixed: table cards alone perform about the same as raw
historical SQL. The best optimized variant retrieves both cards and raw SQL, scoring
$9.00\%$ versus $6.33\%$ (p-value $0.12$) for the comparable baseline on a held-out $N{=}300$ subset,
using retrieved context and harness changes but no agentic loop. 
\end{abstract}

\section{Introduction}

Successful real-world deployments of large language model (LLM) systems depend not just on the
model but also on the information that reaches the model and the harness that manages the model's
interaction with the external world. A growing body of work optimizes this harness--the
executable scaffolding that decides what an LLM application stores, retrieves, and
presents--automatically \citep{lee2026metaharness, hu2024adas}.

In this work we consider the complementary problem of optimizing the {\it context artifacts} that
the model uses for downstream tasks. Concurrent work refines knowledge bases and memory
artifacts after they are built \citep{huang2026deeprefine}, or accumulates them online with
hand-designed update rules \citep{biswal2026agentsm}. We instead ask how to \emph{construct} the
artifacts from raw production traces in the first place. \Cref{app:related} gives an extended
discussion of related work.

We distinguish \emph{raw traces} (prior SQL usage records, excluding held-out
queries), \emph{distilled artifacts} (the reusable SQL reference cards synthesized from
those traces), the \emph{harness} (retrieval and generation scaffolding), and the
\emph{injected context} the model actually sees. This distinction matters because our
claim is not that every useful context item must be summarized first, but that historical
SQL-usage signal should be treated as an optimizable input to the Text-to-SQL system.

Enterprise Text-to-SQL is hard for reasons academic benchmarks rarely capture. Business logic
spans thousands of tables and intermediate views, long-horizon planning remains
brittle even for reasoning models \citep{valmeekam2024lrms}, and accuracy degrades as the
input grows despite nominally large context windows \citep{hsieh2024ruler, bai2024longbench}.
Benchmarks such as Spider \citep{yu2018spider}, BIRD \citep{li2023can}, and even Spider~2.0
\citep{lei2024spider2} evaluate on schemas far simpler than production data lakes. Because no model
can ingest a full enterprise catalog at once, deciding \emph{which} evidence to surface for a given
query becomes the central bottleneck.

We study the Text-to-SQL--specific version of this problem and show, in a production enterprise
setting, that SQL reference-card context artifacts distilled from historical query profiles
outperform prompt/tool harness optimization under the same search procedure. We treat the knowledge-base context the model
consumes as the primary optimization target: we represent SQL queries as a DAG of
sub-problems, which makes
critical subtasks--such as identifying which tables are relevant--independently measurable, and
we distill historical query profiles into reusable SQL reference-card artifacts. Unlike fixed
workload-mining recipes \citep{vaidya2025tailorsql, baek2025kbtext2sql}, our distillation function
is optimized end-to-end against downstream SQL quality using an AlphaEvolve-style search
\citep{alphaevolve2025}.

Our contribution is a distillation-time optimization method that learns to convert
historical data-warehouse queries into reusable SQL reference-card artifacts. To make enterprise
context bottlenecks independently measurable we adopt a query-DAG supervision view--the same
family of intermediates that enterprise benchmarks like BEAVER annotate \citep{chen2024beaver},
here recovered from production SQL rather than hand-labeled--and use it to ablate which subtasks
and intermediate graph information the model is given--table linkage, output schemas, and the
surrounding graph structure--to locate where the bottlenecks actually lie. We then
quantify the trade-off between the two optimization surfaces: on the internal benchmark,
optimizing context artifacts yields larger relative end-to-end AST gains within each model
than prompt/tool harness optimization (${\sim}12\%$ vs.\ ${\sim}3\%$ for Sonnet;
${\sim}25\%$ vs.\ ${\sim}12\%$ for Qwen). Finally, we test
the procedure on BEAVER \citep{chen2024beaver}, a public enterprise SQL benchmark graded by
execution accuracy. Because BEAVER lacks the production-usage signals available internally,
it serves as a conservative transfer check. The result is more nuanced than the internal
benchmark: cards-only and raw-query retrieval are statistically indistinguishable on our held-out
$N{=}300$ subset. The best-scoring held-out variant retrieves both cards and raw SQL,
scoring $9.00\%$ versus $6.33\%$ for the comparable baseline, using a single generation call
and no agentic exploration. This difference is directional rather than statistically significant.

\section{Query-DAG Supervision Framework}
\label{sec:framework}

To measure context quality at the right granularity, we represent each production query as
a directed acyclic graph (DAG) of sub-problems, in the same spirit as the subtask
annotations of enterprise benchmarks like BEAVER~\citep{chen2024beaver}. This scaffold
exposes verifiable intermediates, defines \emph{context-fidelity} controls, and yields
cheap supervision signals that avoid executing arbitrary queries at scale. We use it to
diagnose where context matters (\Cref{sec:diagnosis}), motivating the context-artifact
optimization of \Cref{sec:optimization}. The optimized generator consumes retrieved text
artifacts and raw SQL examples, not predicted query DAGs.

\subsection{Production SQL as a query DAG}
\label{sec:dag}

We represent a production query as a DAG $\cG=(V,E,\cD)$, where each node $v \in V$ is a
logical sub-query (a CTE or subquery), each edge $(v_i,v_j)\in E$ denotes dataflow, and
$\cD=\{(v,d_v,\cS_v)\}_{v\in V}$ annotates each node with a natural-language description
$d_v$ and an output schema $\cS_v$. Each node additionally carries an \emph{input schema}:
the source tables or upstream node outputs it reads. Importantly, this representation is
\emph{recovered from real production SQL} rather than hand-authored: we parse each query
into an abstract syntax tree, extract CTEs and subqueries as nodes, and resolve
column-level lineage to establish edges (\Cref{app:benchmark-details}). The resulting
graphs are an order of magnitude more complex than academic benchmarks--production
queries in our corpus average ${\sim}7$ intermediate steps and reference ${\sim}5$ source
tables, versus the single-digit table counts of Spider~\citep{yu2018spider} and
BIRD~\citep{li2023can}.

\subsection{Verifiable intermediates and graph-fidelity levels}
\label{sec:fidelity}

The payoff of the DAG is that each node is a \emph{verifiable intermediate}: its
description, input linkage, and output schema can each be checked against ground truth
\emph{without executing the full query}--essential in enterprise settings where
end-to-end execution is expensive and governed by data-access controls. This lets us
define a nested hierarchy of graph-fidelity levels, each adding one more slice of the
ground-truth graph to what the model is given:
\begin{alignat*}{4}
\cI_1 &= \{d_v\}_{v \in V}  &&\;\text{(NL-only)} &\qquad \cI_2 &= \cI_1 \cup E  &&\;\text{(+ input linkage)}\\
\cI_3 &= \cI_2 \cup \{\cS_v\}_{v \in V}  &&\;\text{(+ output schemas)} &\qquad \cI_4 &= \cI_3 \cup \{\text{input schemas}\}  &&\;\text{(+ full graph)}
\end{alignat*}
Because $\cI_1 \subset \cI_2 \subset \cI_3 \subset \cI_4$, these levels are the controls we
turn in the diagnosis of \Cref{sec:diagnosis}. 

\subsection{Benchmark and metrics}
\label{sec:benchmark}

We instantiate the framework on an internal benchmark of 5176 production queries drawn
from a large enterprise data warehouse, each paired with its ground-truth SQL, an
LLM-generated natural-language intent, and the source tables and schemas.

We score a generated query against ground truth at the granularity the DAG exposes.
Representing a query as sub-queries $Q=\{q_1,\dots,q_m\}$ with predicted counterparts
$\hat{Q}$, and letting $O_k$ be the number of AST operations in $q_k$ (a complexity
weight), we report
\begin{align}
\text{AST sim.} &= 1 - {\textstyle\sum_k} O_k\, d_{\mathrm{AST}}(q_k,\hat{q}_k) \big/ {\textstyle\sum_k} O_k, \label{eq:ast}\\
\text{String sim.} &= 1 - {\textstyle\sum_k} O_k\, d_{\mathrm{STR}}(q_k,\hat{q}_k) \big/ {\textstyle\sum_k} O_k, \label{eq:str}\\
\text{Linkage sim.} &= |E(Q)\cap E(\hat{Q})| \big/ |E(Q)\cup E(\hat{Q})|, \label{eq:link}
\end{align}
where $d_{\mathrm{AST}}$ and $d_{\mathrm{STR}}$ are normalized edit distances on AST and
string representations and~\eqref{eq:link} is the Jaccard index on DAG edges. We
additionally report an LLM-judge semantic-similarity score (\Cref{app:correlation}) 
and an execution accuracy metric on a subsampled set of queries.

\subsection{Diagnosis: graph fidelity or retrieved content?}
\label{sec:diagnosis}

We now use the framework to ask the question that motivates the rest of the paper: holding
the model fixed, which lever is more valuable--the fidelity of the query \emph{graph} we
hand the model, or the knowledge-base \emph{content} it retrieves? To \emph{upper-bound}
the improvement from better graph structure prediction, we provide an \emph{oracle} ground-truth
graph at several levels of granularity $\cI_1$--$\cI_4$; to isolate the impact of content,
we add retrieval (RAG) on top. \Cref{tab:fidelity} measures synthesis quality for Claude
Sonnet~4.5 and Qwen Coder 3-30B.
Every row carrying the oracle graph ($\cI_2$ onward, including $\cI_4{+}$RAG) is a
diagnostic ceiling, not an attainable system: these rows hand the model ground-truth graph
structure that is unavailable at inference, so they bound--rather than report--what
predicting that structure could buy.
 That high-fidelity intermediates help is not
itself new: on BEAVER, supplying gold subtask annotations increases execution accuracy from
${\sim}10.8\%$ to ${\sim}30.1\%$~\citep{chen2024beaver}, yet leaves a large gap. Our ablation
asks the sharper question of \emph{which} lever--graph fidelity or retrieved content--closes
more of that gap, and how to \emph{produce} the necessary signal without oracle annotation
(\Cref{sec:optimization}).

\begin{table}[t]
\centering
\caption{Synthesis quality versus oracle graph fidelity. Rows are ordered by increasing
input; linkage similarity applies only to graph-level rows.}
\label{tab:fidelity}
\begin{sc}
\footnotesize
\begin{tabular}{llccc}
\toprule
& & \multicolumn{3}{c}{Similarity} \\
\cmidrule{3-5}
Model & Input & AST & String & Linkage \\
\midrule
Claude Sonnet 4.5 & \multirow{2}{*}{$\cI_1$: NL} & 0.100 & 0.303 & -- \\
Qwen Coder 3-30B      &                     & 0.082 & 0.281 & -- \\
\midrule
Claude Sonnet 4.5 & \multirow{2}{*}{$\cI_1$: NL + RAG} & 0.278 & 0.447 & -- \\
Qwen Coder 3-30B      &                           & 0.184 & 0.346 & -- \\
\midrule
Claude Sonnet 4.5 & \multirow{2}{*}{$\cI_2$: NL + Linkage} & 0.248 & 0.468 & 0.735 \\
Qwen Coder 3-30B      &                                          & 0.199 & 0.410 & 0.564 \\
\midrule
Claude Sonnet 4.5 & \multirow{2}{*}{$\cI_3$: + Output Schema} & 0.323 & 0.582 & 0.756 \\
Qwen Coder 3-30B      &                                           & 0.271 & 0.531 & 0.607 \\
\midrule
Claude Sonnet 4.5 & \multirow{2}{*}{$\cI_4$: + Full Graph} & 0.341 & 0.597 & 0.860 \\
Qwen Coder 3-30B      &                                        & 0.291 & 0.551 & 0.708 \\
\midrule
Claude Sonnet 4.5 & \multirow{2}{*}{$\cI_4$ + RAG} & \textbf{0.582} & \textbf{0.750} & \textbf{0.881} \\
Qwen Coder 3-30B      &                                & 0.558 & 0.727 & 0.804 \\
\bottomrule
\end{tabular}
\end{sc}
\end{table}

\Cref{tab:fidelity} separates two effects. \textbf{First, graph fidelity matters.}
Within the oracle-graph series, quality rises monotonically: moving from linkage to the full
graph ($\cI_2\!\to\!\cI_4$) improves AST similarity by ${+}0.093$ for Sonnet and ${+}0.092$
for Qwen. This column-specific trend does not mean a partial oracle graph always beats
ordinary retrieval--for Sonnet, NL+RAG ($0.278$) exceeds $\cI_2$ ($0.248$) and nearly
matches $\cI_4$ ($0.341$). Predicting graph structure is therefore a real source of signal,
but retrieved context is already competitive with weaker oracle graph views.

\textbf{Second, retrieved knowledge-base context has a large impact.} With no graph at all,
retrieval over the knowledge base improves AST similarity over NL-only for both models
($0.100\!\to\!0.278$ for Sonnet and $0.082\!\to\!0.184$ for Qwen). When this context is
added on top of the full graph, it produces the largest jump in the table
($\cI_4\!\to\!\cI_4{+}$RAG, $0.341\!\to\!0.582$ for Sonnet and $0.291\!\to\!0.558$ for
Qwen). The implication is not that graphs are irrelevant; it is that knowledge-base context
can be improved offline, before inference.


\section{Optimizing Context Artifacts}
\label{sec:optimization}

The diagnosis of \Cref{sec:diagnosis} is a comparison of optimization surfaces, not a
dismissal of graph structure. Oracle graph fidelity is valuable, but the ablation shows that
retrieved knowledge-base context contributes a larger marginal gain, including when it is
added on top of the full graph. This makes the retrieved content itself the natural object to
optimize: it can be distilled from historical traces once, indexed, and reused at inference.
The resulting problem is to optimize the content of reusable
\emph{context artifacts}--the knowledge-base material the model retrieves--rather than
treating that content as a fixed input. This focus differs from prompt- and
harness-optimization methods~\citep{khattab2023dspy, khattab2024mipro, yuksekgonul2024textgrad,
lee2026metaharness}, which tune how the system reasons and retrieves while holding the
underlying knowledge fixed, and from fixed workload-mining
recipes~\citep{vaidya2025tailorsql, baek2025kbtext2sql}, which build these artifacts with a
hand-designed pipeline rather than optimizing them end-to-end.

\subsection{Agent System}
\label{sec:agent}

At enterprise scale a catalog holds thousands of tables and millions of lines of SQL, far
exceeding any context window, so the model is wrapped in an agent $A_\theta$ that, given a
natural-language query $l$, retrieves a compact evidence set $\cE = H(l, K) \subseteq K$
before generating. The parameters
$\theta = (K, H, P)$ name three optimizable surfaces:
\begin{itemize}
\item \textbf{Knowledge base $K = \bigcup_t \left(R_t \cup \{K_t\}\right)$} -- for each
  table $t$, historical queries $R_t$ plus a synthesized table summary $K_t$. Its
  \emph{content} is a decision variable, not a fixed input.
\item \textbf{Retrieval and generation harness $H$} -- the tools and retrieval logic that select $\cE$ 
    and generate the SQL.
\item \textbf{Instruction prompt $P$} -- the instructions governing planning and generation.
\end{itemize}
To prevent leakage, the query under evaluation $i$ is excluded at inference, yielding
$K_{-i}$. Given benchmark pairs $\{(l_i, q_i)\}$, we seek
\begin{equation}
\label{eq:agent-opt}
\theta^* = \arg\max_\theta\; \frac{1}{n}\sum_{i=1}^{n} \text{score}\bigl(A_\theta(l_i),\, q_i\bigr),
\end{equation}
where $\text{score}$ is one of the metrics of \Cref{sec:benchmark}. Our focus is the
knowledge-base context $\{K_t\}$; we optimize it separately from the harness $H$ and prompt
$P$. In \Cref{sec:results} we evaluate the two separately-optimized surfaces in combination 
in order to attribute downstream gains to each surface and test whether they compound.

\subsection{The distillation abstraction}
\label{sec:distillation}

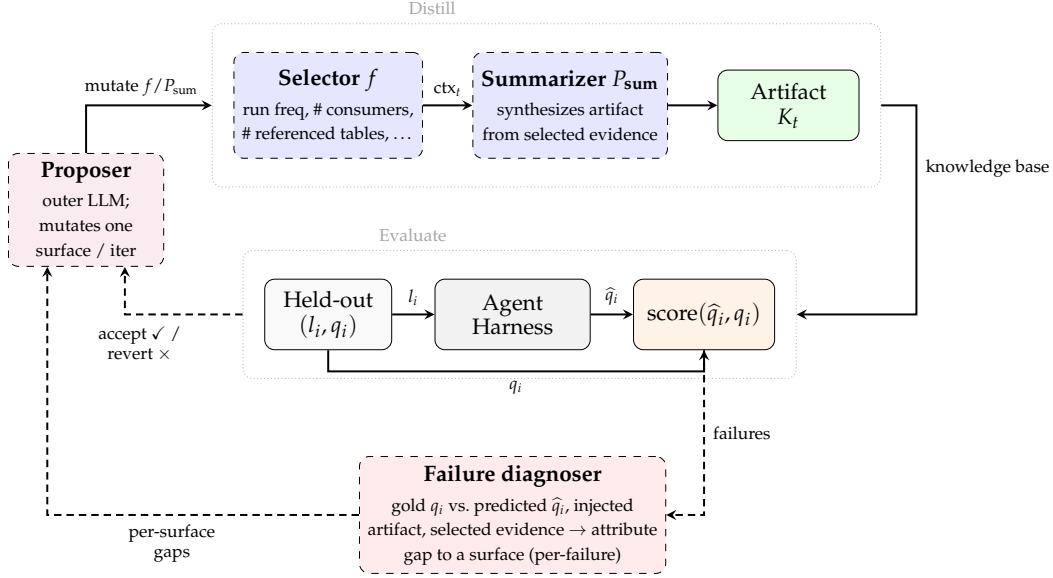
\begin{figure}[t]
\centering
\resizebox{\textwidth}{!}{%
\begin{tikzpicture}[
    node distance=0.8cm and 0.5cm,
    mutable/.style={rectangle, draw, dashed, rounded corners, minimum width=2.6cm, minimum height=1.5cm, align=center, font=\small, fill=#1},
    fixed/.style={rectangle, draw, rounded corners, minimum width=2.0cm, minimum height=1.0cm, align=center, font=\small, fill=#1},
    arr/.style={->, thick, >=stealth},
    feedbackarr/.style={->, thick, >=stealth, densely dashed},
]
    \node[mutable=blue!10] (selector) {
        \textbf{Selector $f$}\\[2pt]
        {\scriptsize run freq, \# consumers,}\\
        {\scriptsize \# referenced tables, \ldots}
    };
    \node[mutable=blue!10, right=0.7cm of selector] (summarizer) {
        \textbf{Summarizer $P_{\text{sum}}$}\\[2pt]
        {\scriptsize synthesizes artifact}\\
        {\scriptsize from selected evidence}
    };
    \node[fixed=green!10, right=0.7cm of summarizer] (readme) {Artifact\\$K_t$};
    \draw[arr] (selector) -- node[above, font=\scriptsize] {$\text{ctx}_t$} (summarizer);
    \draw[arr] (summarizer) -- (readme);
    \node[draw, rounded corners, densely dotted, gray!60,
          fit=(selector)(summarizer)(readme),
          inner xsep=0.3cm, inner ysep=0.4cm,
          label={[font=\scriptsize, text=gray!70]above left:Distill}] (improvebox) {};
    \node[fixed=gray!5, below=1.7cm of selector, minimum width=1.8cm] (heldout) {
        {\footnotesize Held-out}\\{\footnotesize $(l_i, q_i)$}
    };
    \node[fixed=gray!10, right=0.6cm of heldout, minimum width=2.2cm] (harness) {Agent\\Harness};
    \node[fixed=orange!10, right=0.6cm of harness, minimum width=2.0cm] (evalfn) {$\text{score}(\hat{q}_i, q_i)$};
    \draw[arr] (heldout) -- node[above, font=\scriptsize] {$l_i$} (harness);
    \draw[arr] (harness) -- node[above, font=\scriptsize] {$\hat{q}_i$} (evalfn);
    \draw[arr] (heldout.south) -- ++(0,-0.3) -| (evalfn.south)
        node[pos=0.25, below, font=\scriptsize] {$q_i$};
    \node[draw, rounded corners, densely dotted, gray!60,
          fit=(heldout)(harness)(evalfn),
          inner xsep=0.3cm, inner ysep=0.4cm,
          label={[font=\scriptsize, text=gray!70]above left:Evaluate}] (evalbox) {};
    \draw[arr] (improvebox.east) -- ++(0.5,0) |- (evalbox.east)
        node[pos=0.15, right, font=\scriptsize] {knowledge base};
    \node[mutable=red!8, below=1.5cm of harness, minimum width=4.2cm, minimum height=1.3cm] (diag) {
        \textbf{Failure diagnoser}\\[2pt]
        {\scriptsize gold $q_i$ vs.\ predicted $\hat q_i$, injected}\\
        {\scriptsize artifact, selected evidence $\to$ attribute}\\
        {\scriptsize gap to a surface (per-failure)}
    };
    \draw[feedbackarr] (evalfn.south) |- (diag.east)
        node[pos=0.25, right, font=\scriptsize] {failures};
    \node[mutable=purple!8, minimum width=2.2cm, minimum height=1.4cm]
        (proposer) at ($(improvebox.west)!0.5!(evalbox.west)+(-2.0,0)$) {
        \textbf{Proposer}\\[2pt]
        {\scriptsize outer LLM;}\\
        {\scriptsize mutates one}\\
        {\scriptsize surface / iter}
    };
    \draw[arr] (proposer.north) |- (improvebox.west)
        node[pos=0.72, above, font=\scriptsize] {mutate $f$/$P_{\text{sum}}$};
    \draw[feedbackarr] (evalbox.west) -- ++(-0.35,0) -| ([xshift=0.55cm]proposer.south)
        node[pos=0.42, below=1pt, font=\scriptsize, align=center] {accept $\checkmark$ /\\revert $\times$};
    \draw[feedbackarr] (diag.west) -| ([xshift=-0.55cm]proposer.south)
        node[pos=0.3, below, font=\scriptsize, align=center] {per-surface\\gaps};
\end{tikzpicture}
}
\caption{Context-artifact optimization loop. An outer LLM mutates the selector $f$ or
summarizer prompt $P_{\text{sum}}$, regenerates per-table artifacts $K_t$, scores held-out
queries, and accepts or reverts by metric delta. Failure feedback attributes errors to
$f$, $P_{\text{sum}}$, or harness $H$ to target later mutations.}
\label{fig:kb_loop}
\end{figure}

The core object is a \emph{distillation function} that turns raw usage signals into a
compact, reusable context artifact. For each table $t$, a \emph{selector}
$f(R^{-i}_t) \to \text{ctx}_t$ filters that table's queries (excluding query $i$) by
attributes such as run frequency, number of consumers, and number of referenced tables, producing a
compact set of selected evidence $\text{ctx}_t$; a \emph{summarizer}
$\mathrm{LLM}(\text{ctx}_t;\, P_{\text{sum}}) \to K_t^{-i}$ then synthesizes the table's
artifact under prompt $P_{\text{sum}}$. The optimized summaries then augment $K$, improving the evidence 
available to the agent (\Cref{fig:kb_loop}).

This abstraction is deliberately source-agnostic. Internally, the distillation input is raw
production SQL; recovered DAG annotations construct labels and diagnose failures, not
inference artifacts. On BEAVER (\Cref{sec:external}), benchmark-provided SQL, schema
metadata, and BEAVER's own intermediate annotations serve as a weaker public proxy for usage
and are distilled into aggregated table cards. The method does not require proprietary
traces, but its strongest setting is one with a real corpus of prior warehouse usage to
distill from.

\subsection{Optimization procedure}
\label{sec:procedure}

We search over these surfaces with an AlphaEvolve-style autoresearch
loop~\citep{alphaevolve2025}: an outer loop proposes a mutation to a single surface,
scores it on the benchmark, and accepts or reverts based on the metric delta
(\Cref{alg:kb_opt}). The search procedure itself is off-the-shelf; our
contribution is not a new optimizer but what we optimize--the
context-artifact distillation function $(f, P_{\text{sum}})$ that turns raw historical
traces into reusable retrieval artifacts. We place both the selector $f$ and the summarizer
prompt $P_{\text{sum}}$ in the search space and search against downstream SQL quality, rather
than hand-designing this function or holding it fixed. For harness and prompt optimization the mutation targets $H$ or $P$;
for context-artifact optimization it targets the selector $f$ or the summarizer prompt
$P_{\text{sum}}$, a two-level search over \emph{what} goes into each artifact and
\emph{how} it is written. Each search restricts mutations to one surface family
($\{f,P_{\text{sum}}\}$ or $\{H,P\}$); the combined configuration of \Cref{sec:results}
stacks the two separately-optimized surfaces rather than searching them jointly.
On the internal benchmark, the outer proposer evaluates mutations on a 100-query inner-loop
sample spanning 78 referenced tables; accepted artifact mutations regenerate those tables'
summaries with a 8192-token cap. We use Claude Sonnet~4.6 and Qwen Coder 3-30B. 
Each surface search is run until it plateaus.

\begin{algorithm}[t]
\caption{Context-artifact optimization}
\label{alg:kb_opt}
\begin{algorithmic}[1]
\Input Profile corpus $\{R_t\}$, eval set $\mathcal{D}=\{(l_i,q_i,T_i)\}_{i=1}^n$ with relevant tables $T_i$, initial selector $f$ and prompt $P_{\text{sum}}$
\Output Optimized $f^*,\, P^*_{\text{sum}}$
\State $f^*,\, P^*_{\text{sum}} \leftarrow f,\, P_{\text{sum}}$;\quad $\text{best\_score} \leftarrow -\infty$
\Repeat
    \State Mutate one surface in $\theta$ (one per iteration)
    \For{each held-out example $i \in \{1,\ldots,n\}$}
        \For{each relevant table $t \in T_i$}
            \State $\text{ctx}_{t,i} \leftarrow f(R^{-i}_t)$;\quad $K_t^{-i} \leftarrow \mathrm{LLM}(\text{ctx}_{t,i},\, P_{\text{sum}})$
            \State Embed $K_t^{-i}$ into the knowledge base for example $i$
        \EndFor
    \EndFor
    \State $s \leftarrow \text{score}(\mathcal{D},\, \{K_t^{-i}\})$
    \If{$s > \text{best\_score}$} \State accept; $\text{best\_score}\leftarrow s$;\quad $f^*,\, P^*_{\text{sum}} \leftarrow f,\, P_{\text{sum}}$
    \Else\ \State revert \EndIf
\Until{convergence}
\end{algorithmic}
\end{algorithm}

To keep artifact search tractable, candidate artifacts are rebuilt only for relevant
tables known during search; all reported results in \Cref{sec:results} are scored
end-to-end with real retrieval.

\subsection{Error feedback: diagnosing failures to target mutations}
\label{sec:error-feedback}

A scalar score delta tells the outer loop \emph{whether} a mutation helped but not
\emph{why} a candidate still fails, so a search driven by the delta alone mutates its
allowed surface without knowing which part is responsible. We close this gap with an
\textbf{error-feedback} mechanism (\Cref{fig:kb_loop}). After a candidate is scored, a
\emph{failure diagnoser}--an LLM call--examines each wrong prediction together with the
artifacts that produced it: the natural-language question, the gold query $q_i$, the
model's prediction $\hat q_i$, the \emph{injected artifact} the model actually saw, and the
\emph{selected evidence} that fed the summarizer. For each failure it attributes the
missing information to exactly one optimizable surface:
(1) \textbf{Summarizer ($P_{\text{sum}}$) gap} -- the needed fact \emph{was} present in
  the selected evidence but did not survive into the artifact; the summarizer prompt should
  surface it.
(2) \textbf{Selector ($f$) gap} -- the needed fact was in \emph{none} of the selected
  evidence; the selector should expose different or additional usage signals.
(3) \textbf{Harness ($H$) gap} -- the artifact was adequate but the generation step errored
  for a prompt-fixable reason (e.g.\ wrong SQL dialect, or output not emitted in the
  required form); the harness instruction should constrain it.

Aggregating these attributions across the failed tasks yields a per-surface ranking of
\emph{missing-information categories} (each with a suggested fix),
which the outer loop can use to target the next mutation at the surface most responsible for
the residual errors. Note that the diagnoser reads the concrete failing examples but its
\emph{output}--the categories and suggested fixes--is generic, so the optimizer is
steered toward better extraction \emph{procedures} without copying instance-specific values
into a prompt.

\section{Empirics}
\label{sec:results}

We now test whether \emph{automatically generated} context artifacts deliver the gains
suggested by the oracle-graph diagnosis (\Cref{sec:diagnosis}), and how they compare to
optimizing the harness.

\paragraph{Metrics.}
On the internal benchmark we report AST, string, and linkage similarities
(\Cref{sec:benchmark}), LLM-judge semantic similarity (\Cref{app:correlation}), and
table-selection recall/precision. Because arbitrary production queries cannot be re-executed
at scale--most read tables the evaluation cluster is not granted access to, or depend on
upstream state we cannot reconstruct--these structural metrics are the primary internal
signals, and we use BEAVER as a public execution-graded check.

We additionally report execution accuracy (EX) on a separate cohort of $n{=}102$ 
production queries, becase not every query in our corpus was executable in our test environment. 
Only $11$ of the $102$ also appear in the $517$-query sample.

\subsection{Internal Production Text2SQL Benchmark}
\label{sec:internal}

We compare four configurations: baseline versus optimized retrieval harness, crossed with
baseline versus optimized knowledge-base context. Every configuration is evaluated with the
same end-to-end retrieval-and-generation protocol. \Cref{tab:main} ablates the 
two optimization surfaces--harness ($H$, $P$) and knowledge-base
context ($K$)--against the baseline, for both models.

\paragraph{Benchmark, models, and retrieval.}
We evaluate on the internal benchmark of \Cref{sec:benchmark} (5176 production queries,
${\sim}100$K-profile corpus) with two models--Claude Sonnet~4.6 and Qwen Coder
3-30B--so conclusions are not tied to a single model family. All results use real
retrieval at test time: no oracle table linkage is supplied to the generator. The oracle
linkage used inside \Cref{sec:procedure} only reduces the cost of constructing candidate
artifacts during search. 

\paragraph{Baselines.}
The all-baseline cell uses two deliberately simple defaults. The \textbf{context baseline}
is handcrafted table documentation plus raw query profiles $R_t$; context optimization
replaces the documentation with distilled SQL reference cards (\Cref{sec:optimization}),
and \Cref{app:cards} shows a representative card. The \textbf{harness baseline} is a
single vector-search retrieval tool with $k{=}10$ and a hand-written generation prompt;
harness optimization searches over retrieval tools $H$ and prompt $P$ (\Cref{app:prompts}).
Each row in \Cref{tab:main} names which surfaces are optimized, with the all-baseline row as
the common reference point.

\begin{table}[t]
\centering
\caption{Internal benchmark results: harness optimization versus context-artifact
optimization, evaluated end-to-end with real retrieval. Table-selection and similarity
metrics use the $n{=}517$ sample; EX uses the executable cohort with $n{=}102$. ``Context''
= table schema information vs.\ optimized SQL reference cards.}
\label{tab:main}
\footnotesize
\begin{sc}
\resizebox{\textwidth}{!}{%
\begin{tabular}{lllcccccc}
\toprule
& & & \multicolumn{2}{c}{Table Selection} & \multicolumn{3}{c}{End-to-End} & Exec. \\
\cmidrule(lr){4-5}\cmidrule(lr){6-8}\cmidrule(lr){9-9}
Model & Harness & Context & Recall & Prec. & AST & String & LLM-J & EX \\
\midrule
\multirow{4}{*}{Claude Sonnet 4.6}
 & Baseline  & Baseline  & 0.714 & 0.658 & 0.490 & 0.591 & 0.553 & 0.255 \\
 & Optimized & Baseline  & 0.741 & 0.679 & 0.503 & 0.600 & 0.546 & 0.275 \\
 & Baseline  & Optimized & \textbf{0.805} & \textbf{0.777} & 0.550 & 0.639 & 0.597 & \textbf{0.333} \\
 & Optimized & Optimized & 0.766 & 0.717 & \textbf{0.570} & \textbf{0.655} & \textbf{0.600} & 0.304 \\
\midrule
\multirow{4}{*}{Qwen Coder 3-30B}
 & Baseline  & Baseline  & 0.660 & 0.577 & 0.407 & 0.508 & 0.503 & 0.176 \\
 & Optimized & Baseline  & 0.726 & 0.621 & 0.456 & 0.547 & 0.516 & 0.206 \\
 & Baseline  & Optimized & 0.737 & \textbf{0.687} & 0.509 & 0.594 & 0.550 & 0.235 \\
 & Optimized & Optimized & \textbf{0.740} & 0.631 & \textbf{0.519} & \textbf{0.603} & \textbf{0.566} & \textbf{0.284} \\
\bottomrule
\end{tabular}}%
\end{sc}
\end{table}

Optimizing the prompt and retrieval tools over the baseline knowledge base improves table
selection (recall ${+}4\%$ for Sonnet, ${+}10\%$ for Qwen) and end-to-end AST similarity
(${+}3\%$ and ${+}12\%$). The accepted harness changes mostly affect retrieval: the best
configuration issues two calls (query search and table-documentation search) over both
semantic and keyword indices, and an evidence-voting variant best serves table selection.

Holding the harness at baseline and replacing documentation with optimized
SQL reference cards improves AST similarity by ${\sim}12\%$ relative for Sonnet and
${\sim}25\%$ relative for Qwen. In contrast to the BEAVER execution estimates below, this
internal structural effect is resolved at $n{=}517$: for Sonnet the $95\%$
intervals for baseline
documentation and optimized cards are separated
($0.490\,[0.461,0.520]\!\to\!0.550\,[0.521,0.579]$), and the Qwen optimized-cards interval
is comparably tight ($0.509\,[0.479,0.541]$). The effect is larger than the harness-only relative gain within
each model (${\sim}3\%$ for Sonnet and ${\sim}12\%$ for Qwen); under the same
retrieval/generation harness, table-selection recall rises to $0.81$ / $0.74$. The
optimized artifact is a SQL reference-card consisting of: verbatim SQL fragments, join recipes,
filter templates, and example CTEs, with roughly $60\%$ of the token budget spent on
concrete SQL examples. In this internal ablation, compact optimized evidence outperforms
baseline documentation under the same harness.

Combining both optimizations roughly matches--but does not exceed--the better single
surface. On Qwen the harness adds ${\sim}12\%$ AST similarity over baseline documentation but only
${\sim}2\%$ on top of optimized cards; on Sonnet the combination is marginally best on
AST/string-similarity and LLM-judge, but is within the CIs of the artifacts-only
configuration. The tweo surfaces partially substitute: once the
right content is in front of the model, smarter retrieval has less to recover. On the
executable cohort where we measure execution accuracy (EX), the estimates move in the same
direction: the content-only rows have higher EX point estimates by ${+}0.078$ on Sonnet
($0.255\!\to\!0.333$) and ${+}0.059$ on Qwen ($0.176\!\to\!0.235$), versus ${+}0.020$ and
${+}0.030$ for the harness. Because the executable cohort is small, we read EX as
directional corroboration: at $n{=}102$ each CI is roughly $\pm 0.08$ wide and
every EX interval overlaps the others within its model block (\Cref{tab:main-ci}). Despite
not being statistically significant, the execution accuracy improvements align directionally with
the structural metrics. 

\paragraph{What errors remain?}
A failure analysis (\Cref{app:failure}) shows that optimization reduces some
retrieval-phase errors, but persistent schema- and instance-linking failures remain.

\subsection{External validation: BEAVER Benchmark}
\label{sec:external}

Our internal benchmark is proprietary and scored primarily by structural proxies, so we turn to
BEAVER \citep{chen2024beaver}--a public enterprise Text-to-SQL benchmark drawn from real
private data warehouses and graded by \emph{execution accuracy}--to test how far
context-artifact distillation transfers beyond our internal benchmark.
We evaluate on a fixed $N{=}300$ subset drawn once (seed $20260617$) from BEAVER's
$5787$-query \texttt{dw} development split, stratified by query compositionality and kept
fixed across conditions. Each question is answered with the same evaluation scaffold:
dense retrieval over a per-table context index, generation with Claude Sonnet~4.5,
execution against BEAVER's MySQL database, and BEAVER's official set-based scoring. 
No oracle table linkage is given. Published rows in \Cref{tab:beaver} are external
reference points only; optimized rows remove held-out task IDs before building raw-query
or table-card indices.


\paragraph{Experiment Setup.} We ask whether \emph{distilling} usage into reference cards
adds anything over retrieving the raw historical SQL traces directly. We compare three
optimized context channels:
(1) \textbf{Aggregated table cards} -- one card per table, distilling many non-held-out
  historical queries for that table into join recipes, filter idioms, and a representative example.
(2) \textbf{Raw historical queries} -- the retrieved gold SQL of non-held-out questions
  referencing the table, injected verbatim with no distillation.
(3) \textbf{Both} -- retrieve cards and raw queries separately, inject both.
In each setting, we optimize the harness and, in the case of table cards, the context.

\paragraph{Optimization Procedure.} We select configurations on disjoint development splits
using the optimization loop of \Cref{sec:procedure,sec:error-feedback}, then evaluate the
selected configurations once on the held-out $N{=}300$ test subset (\Cref{tab:beaver}).
Development accuracy is higher than held-out test accuracy, as expected when the search
selects on development folds.

\begin{table}[t]
\centering
\caption{External validation on BEAVER (execution accuracy, held-out $N{=}300$; no oracle
table linkage). Exec.\ acc.\ is reported with a 95\% CI over the 300 binary
outcomes. Published rows ($\dagger$) are point estimates from the benchmark authors'
harness~\citep{chen2024beaver}. Pairwise differences among the optimized rows are not significant 
under paired t-tests: Both vs.\ raw $p{=}0.12$, Both vs.\ cards $p{=}0.14$.}
\label{tab:beaver}
\footnotesize
\begin{sc}
\resizebox{\textwidth}{!}{%
\begin{tabular}{llccc}
\toprule
Method & Context injected & Tools / calls & Exec.\ acc. & 95\% CI \\
\midrule
Few-shot (ours) & schemas + demos      & --             & 6.33\% & {\footnotesize [4.1, 9.7]}  \\
Few-shot$^{\dagger}$       & schemas + demos             & --              & 8.8\%  & -- \\
ReFoRCE$^{\dagger}$        & self-explored schema        & explore, vote, fix & 10.8\% & -- \\
\midrule
Raw queries (Optimized) & retrieved gold SQL        & --          & 6.33\% & {\footnotesize [4.1, 9.7]}  \\
Table cards (Optimized) & aggregated per-table      & -- & 6.67\% & {\footnotesize [4.4, 10.1]}  \\
Both (Optimized) & cards + raw queries   & -- & \textbf{9.00\%} & {\footnotesize [6.3, 12.8]} \\
\bottomrule
\end{tabular}}%
\end{sc}
\end{table}

\paragraph{Results} On this held-out subset, the optimized cards+raw system scores
$9.00\%$, compared with $6.33\%$ for our directly comparable pre-optimization harness:
a $+2.67$ point difference ($\sim$$42\%$ relative) on the same 300 questions, with the
same generator, one generation call, and no agentic loop. At $N{=}300$, this is a
directional result with a p-value of p=0.12 on the paired test. 

The content-source comparison is
nuanced: cards alone are similar to raw-query retrieval ($6.67\%$ vs.\ $6.33\%$;
discordant pairs split near-evenly, $11/10$), while cards+raw gives the best score
($27/300$). Since that arm receives more total context, we treat BEAVER as an encouraging
but non-decisive transfer check: the combined context scores highest, while the
cards-only comparison is tied with raw SQL at this sample size. This smaller effect size on BEAVER
 is not surprising as BEAVER lacks the production-usage signals our method is able
to exploit. Its table context is limited to column identifiers, column types, and a few example rows,
and SQL queries without usage signals.


\paragraph{Limitations.} There are several limitations of our work. \emph{(i) Scoring.} The internal
benchmark is graded primarily by structural proxies (AST, string, and linkage similarity)
and an LLM judge; execution accuracy is available on a smaller executable cohort
($n{=}102$), because most production profiles cannot be re-executed at scale. \emph{(ii)
Statistical power.} The public execution-graded result uses an $N{=}300$ BEAVER subset that
is underpowered for the small effect sizes we observe, so every BEAVER comparison here
is directional (\Cref{tab:beaver}). \emph{(iii) Compositionality.} The gains concentrate on
table selection and schema grounding; deeply compositional queries stay near zero
regardless of injected content (\Cref{tab:fail-beaver}), leaving query structure as a
separate bottleneck. \emph{(iv) Regime.} We study a single-call,
retrieval-only harness rather than a multi-turn or RL-trained agent. 

\subsubsection*{LLM Usage Disclosure}
We used LLMs in this work -- in addition to human effort -- to perform more extensive literature reviews,
implement code with tools like CoPilot, and to edit the writing of this paper.

\clearpage

\bibliographystyle{styles/ims_nourl_eprint}
\bibliography{biblio,internal,external}
\clearpage

\appendix
\crefalias{section}{appendix}
\crefalias{subsection}{appendix}
\crefalias{subsubsection}{appendix}

\section{Extended related work}
\label{app:related}

A large body of Text-to-SQL work improves schema-aware SQL generation: representing the
question and database schema, linking mentions to tables and columns, and constraining the
query produced by the model. Schema-aware parsers such as
RAT-SQL~\citep{wang2019ratsql} and IRNet~\citep{guo2019irnet} make schema structure and
schema linking explicit, while RESDSQL~\citep{li2023resdsql} separates schema linking from
SQL-skeleton prediction. LLM-era systems move more of this structure into prompting and
control: DIN-SQL~\citep{pourreza2023dinsql} decomposes generation into smaller
in-context subproblems, DAIL-SQL~\citep{gao2023dail_sql} studies how examples should be
selected and organized at inference time, PICARD~\citep{scholak2021picard} constrains decoding with an
incremental parser, SkyRL-SQL~\citep{liu2025skyrlsql} trains a multi-turn agent to probe
databases, refine queries, and verify results, and ReFoRCE~\citep{deng2025reforce}
combines schema compression, self-refinement, consensus, and column exploration. These
methods primarily target representation, decoding, prompting, or interaction at generation
time. Our focus is complementary: deciding what reusable evidence should exist in the
knowledge base before retrieval and generation begin.

The closest Text-to-SQL work uses historical workload signal. TailorSQL~\citep{vaidya2025tailorsql}
exploits past queries because they reveal common join paths and obscure schema semantics
that are absent from table names alone. \citet{baek2025kbtext2sql} construct a reusable
knowledge base from available questions, schemas, and associated knowledge, improving
knowledge-augmented Text-to-SQL across datasets. AgentSM~\citep{biswal2026agentsm}
stores prior execution traces as structured semantic memories that guide future agent
trajectories, and DeepRefine~\citep{huang2026deeprefine} refines an already constructed
agent-compiled knowledge base through multi-turn diagnosis and targeted refinement.
Together, this line of work shows that external artifacts and past traces are useful system
inputs, rather than static documentation. The key difference is that we search over how
SQL-workload artifacts are constructed from raw traces, using downstream SQL quality as the
acceptance criterion. The search can change the format, content, and granularity of the
reference cards, rather than assuming a fixed workload-mining template or refining only an
already constructed knowledge base.

Our optimization procedure is also related to work that treats prompts, programs, and agent
harnesses as learnable objects. DSPy~\citep{khattab2023dspy} abstracts LM pipelines as
parameterized computational graphs; MIPRO~\citep{khattab2024mipro} optimizes instructions
and demonstrations for multi-stage LM programs; and TextGrad~\citep{yuksekgonul2024textgrad}
uses natural-language feedback as a gradient-like signal through compound AI systems.
AlphaEvolve~\citep{alphaevolve2025} and ADAS~\citep{hu2024adas} search over code or agent
designs with evaluator feedback, while Meta-Harness~\citep{lee2026metaharness} searches
over the harness code that stores, retrieves, and presents information to the model. We use
a similar outer-loop accept/revert search, but make the knowledge content an optimized
surface alongside the retrieval harness and prompt. Keeping these surfaces separate is what
allows the attribution in \Cref{sec:results}: in our setting, changing the artifact content
can dominate changing the agent scaffolding around it.

The evaluation setting is shaped by the gap between public benchmarks and production data
lakes. Spider~\citep{yu2018spider} introduced cross-domain generalization, and
BIRD~\citep{li2023can} added larger databases, external knowledge, and database-value
grounding. Spider~2.0~\citep{lei2024spider2} and BEAVER~\citep{chen2024beaver} move
closer to enterprise use cases with larger schemas, realistic workflows, and domain
knowledge requirements. Our internal benchmark pushes on the same regime but uses private
production traces, so arbitrary execution is costly and often impossible. This motivates
the structural metrics of \Cref{sec:benchmark} and the external BEAVER validation in
\Cref{sec:external}.

Finally, Text-to-SQL evaluation itself is imperfect. Execution can be useful as a decoding
or validation signal~\citep{wang2018executionguided}, but benchmark-level execution
accuracy can penalize semantically valid alternatives, reward structurally wrong queries
that happen to match outputs, and change model rankings under closer inspection
\citep{pourreza2023evaluate, ascoli2024etm, kim2025flex}. These concerns are especially
acute for long enterprise queries, where partial equivalence, dialect behavior, and
underspecified natural-language requests are common. We therefore report structural and
judge-based proxies on the internal benchmark, and use execution-graded BEAVER as an
external check that the main conclusion is not an artifact of proxy scoring.

\section{Benchmark construction details}
\label{app:benchmark-details}

The internal benchmark is derived from recurring query profiles logged by an enterprise
workload-orchestration system. Each profile contains production SQL and usage metadata. From
the ${\sim}100$K-profile corpus we select 5176 queries with three filters:
\begin{enumerate}
\item \textbf{Relevance:} retain only queries that reference a table in the target business
  domain, the use case we evaluate against.
\item \textbf{Version control:} keep only the most recent version of each profile, avoiding
  near-duplicate revisions of the same query.
\item \textbf{Execution validation:} keep only queries that executed successfully within the
  past three years, as a coarse proxy for production viability.
\end{enumerate}

\paragraph{Canonicalization.}
Each SQL string is normalized before DAG extraction: we parse with SQLGlot, normalize
identifiers and aliases, standardize intermediate-table creation to \verb|CREATE TEMP TABLE|,
and flatten nested sub-query expressions inside \verb|CREATE| statements. These transforms
produce a uniform representation for structural comparison; they are not intended to change
query semantics.

\paragraph{Query-profile DAG.}
On the canonicalized AST we run lineage analysis to build the query DAG of \Cref{sec:dag}.
Source tables form the input nodes, temporary tables and CTEs form internal nodes, and the
final \verb|SELECT| is the sink. Column-level lineage defines the edges. The resulting graph
is the basis for the verifiable intermediates and fidelity levels in
\Cref{sec:fidelity}.

\paragraph{Ground-truth labels via LLM annotation.}
We use Qwen Coder 3-30B to annotate each DAG node with a natural-language sub-problem
description and each full profile with a natural-language intent. The gold SQL remains the
original human-authored production query; only the natural-language labels are model
generated. This avoids relying on stale analyst-written descriptions while keeping the SQL
target fixed.

\paragraph{Graph-structure input diagnosis.}
The annotated DAG lets us construct, for each query, the various inputs
ablated in \Cref{sec:diagnosis}: the natural-language intent and target output schema
(\emph{NL}); the script-level input tables and their schemas (\emph{linkage}); per-node
natural-language descriptions of the ground-truth graph (\emph{NL~+~linkage}, $\cI_2$); the
per-node output schemas ($\cI_3$); and the full set of inter-node linkages including
node-level input/output schemas (full graph, $\cI_4$). Supplying successively more of this
ground-truth structure is exactly the oracle ablation of \Cref{tab:fidelity}.

\section{Confidence intervals for the internal benchmark}
\label{app:cis}

\Cref{tab:main} reports point estimates; \Cref{tab:main-ci} gives a $95\%$ interval for every
one of those cells. For the continuous metrics--table-selection recall and precision, and the
three end-to-end similarities--these are bootstrap percentile intervals computed over the
per-task scores at evaluation time ($2000$ resamples, seed $42$). 

\begin{table}[h]
\centering
\caption{$95\%$ confidence intervals for every cell of \Cref{tab:main}. Effective per-cell $n$ after evaluation failures is
$488$--$517$ for table selection and $509$--$517$ end-to-end; EX is $n{=}102$ in every cell.
``Context'' baseline = handcrafted documentation, optimized = SQL reference cards.}
\label{tab:main-ci}
\footnotesize
\begin{sc}
\resizebox{\textwidth}{!}{%
\begin{tabular}{lllcccccc}
\toprule
& & & \multicolumn{2}{c}{Table Selection} & \multicolumn{3}{c}{End-to-End} & Exec. \\
\cmidrule(lr){4-5}\cmidrule(lr){6-8}\cmidrule(lr){9-9}
Model & Harness & Context & Recall & Prec. & AST & String & LLM-J & EX \\
\midrule
\multirow{4}{*}{Claude Sonnet 4.6}
 & Baseline  & Baseline  & [0.682, 0.745] & [0.628, 0.689] & [0.461, 0.520] & [0.564, 0.618] & [0.525, 0.580] & [0.180, 0.347] \\
 & Optimized & Baseline  & [0.712, 0.770] & [0.650, 0.709] & [0.473, 0.533] & [0.574, 0.628] & [0.518, 0.574] & [0.197, 0.368] \\
 & Baseline  & Optimized & [0.778, 0.830] & [0.749, 0.802] & [0.521, 0.579] & [0.613, 0.665] & [0.570, 0.625] & [0.249, 0.429] \\
 & Optimized & Optimized & [0.738, 0.793] & [0.689, 0.745] & [0.540, 0.599] & [0.627, 0.681] & [0.571, 0.629] & [0.223, 0.399] \\
\midrule
\multirow{4}{*}{Qwen Coder 3-30B}
 & Baseline  & Baseline  & [0.628, 0.690] & [0.545, 0.607] & [0.381, 0.435] & [0.484, 0.534] & [0.478, 0.531] & [0.115, 0.262] \\
 & Optimized & Baseline  & [0.698, 0.756] & [0.593, 0.651] & [0.424, 0.486] & [0.519, 0.575] & [0.487, 0.544] & [0.139, 0.294] \\
 & Baseline  & Optimized & [0.710, 0.763] & [0.658, 0.714] & [0.479, 0.541] & [0.567, 0.622] & [0.521, 0.578] & [0.164, 0.326] \\
 & Optimized & Optimized & [0.712, 0.768] & [0.601, 0.661] & [0.490, 0.549] & [0.577, 0.630] & [0.537, 0.594] & [0.206, 0.378] \\
\bottomrule
\end{tabular}}%
\end{sc}
\end{table}

From \cref{tab:main-ci} we see that holding the harness at baseline, the
Sonnet AST intervals for baseline documentation and optimized cards do not overlap
($[0.461, 0.520]$ vs.\ $[0.521, 0.579]$, though only barely), and the table-selection recall
intervals separate more comfortably for both models. 
For the EX column, at $n{=}102$ each CI spans roughly $\pm 0.08$, and within each model
block every EX interval overlaps every other one--so the EX ordering is consistent with the
structural metrics but cannot on its own establish that ranking. The same caveat applies to
the combined configuration, whose intervals overlap those of the better single surface on
every metric.

\section{Harness and prompt optimization: search space and accepted mutations}
\label{app:prompts}

The optimized-harness column of \Cref{tab:main} comes from the same autoresearch loop as
the artifact search, but with mutations restricted to retrieval tools $H$ and instructions
$P$. The accepted retrieval changes increase independent evidence per call: the baseline is
a single semantic search over the indexed corpora ($k{=}10$), while the best end-to-end
configuration searches both dense and keyword indices over profiles and documentation. For
table selection, the best mutation uses evidence voting: it dispatches a fixed slate of
diverse searches and ranks candidate tables by how many searches surfaced them.

Prompt mutations were smaller. The accepted SQL-generation edits emphasize request
coverage and forbid guessing columns not present in retrieved evidence. The accepted
table-selection edits shift the model away from a strict precision-only rule toward a
recall-weighted rule: select only grounded candidates, but prefer one uncertain plausible
source table over dropping a required table. These changes explain the table-selection
recall gains in \Cref{tab:main}; they are less important than the retrieval-tool changes.

\section{LLM-judge versus AST-similarity correlation}
\label{app:correlation}

\begin{figure}[t]
\centering
\includegraphics[width=0.66\textwidth]{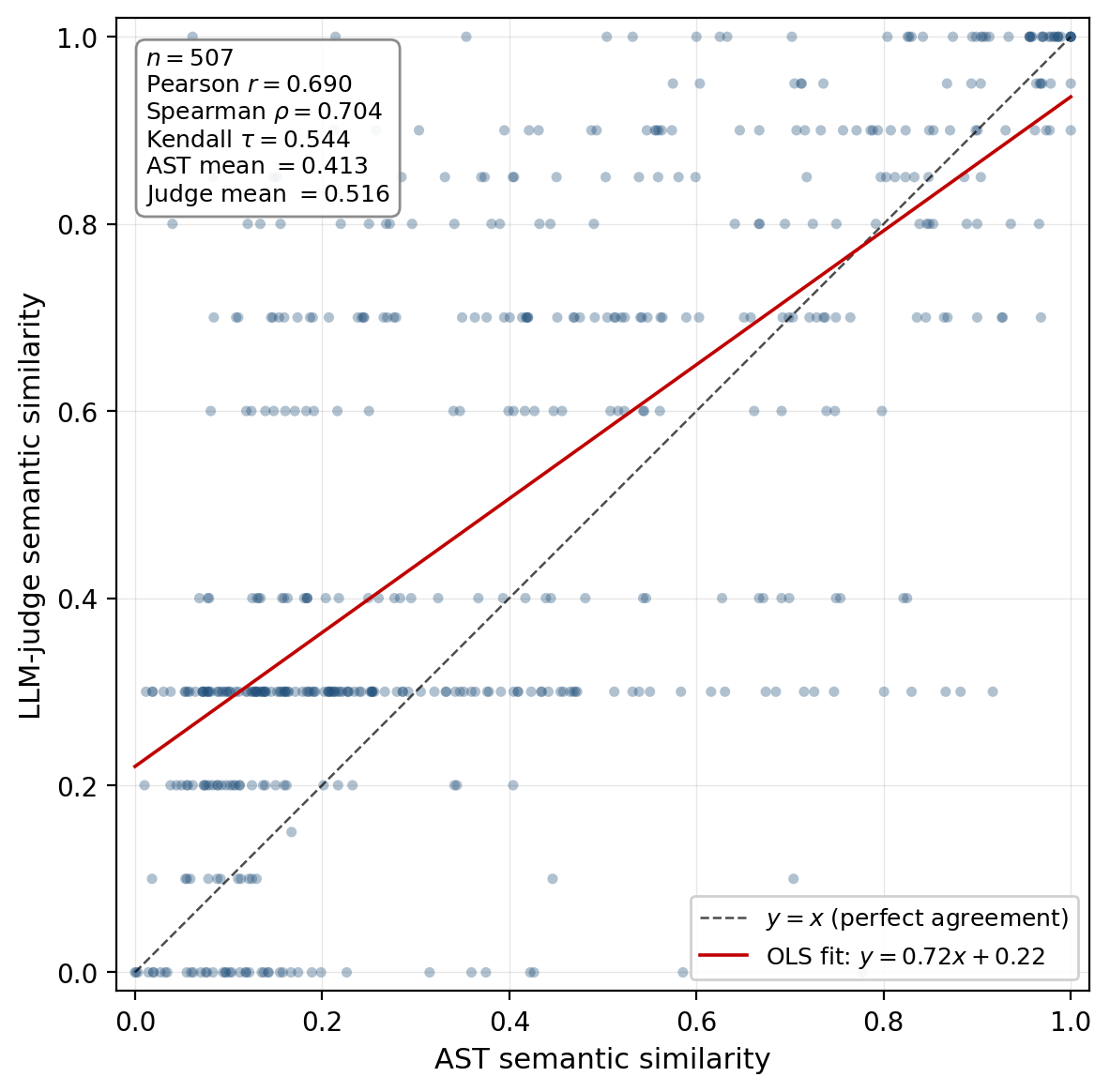}
\caption{LLM-judge semantic similarity versus AST similarity on the internal benchmark. The
positive association supports using the judge as a complementary semantic signal to the
structural metrics.}
\label{fig:ast-vs-llm-judge}
\end{figure}

The LLM judge of \Cref{sec:results} (Qwen 3 Coder Next) scores whether the generated and
gold SQL would answer the same request, ignoring formatting, wrapper statements, and
semantically equivalent rewrites. Judge score and AST similarity move together
(\Cref{fig:ast-vs-llm-judge}): the judge preserves the ordering of configurations while
crediting semantically equivalent queries that differ syntactically, which is why we report
it alongside the structural proxies.

\section{Our few-shot baseline versus the published BEAVER baseline}
\label{app:beaverref}

Our reproduction of BEAVER's few-shot baseline obtains $6.33\%$ ($19/300$), while the benchmark
authors report $8.8\%$ for the same generator family in their setting~0 (no oracle tables). These
two numbers are measured on different question samples: our figure is the $300$-question
stratified draw of \Cref{sec:external}, while the published cell is measured on the release's own
subsample of the \texttt{dw} development split, which the download script regenerates from a fixed
seed at install time rather than shipping. Because the exact question sets are not aligned, this
comparison is descriptive rather than paired, and the published cell is useful as context rather
than as a direct target; $8.8\%$ lies inside the $95\%$ confidence interval $[4.1, 9.7]$ of our own
$19/300$.

\section{Failure analysis}
\label{app:failure}

\begin{figure}[t]
\centering
\includegraphics[width=\textwidth]{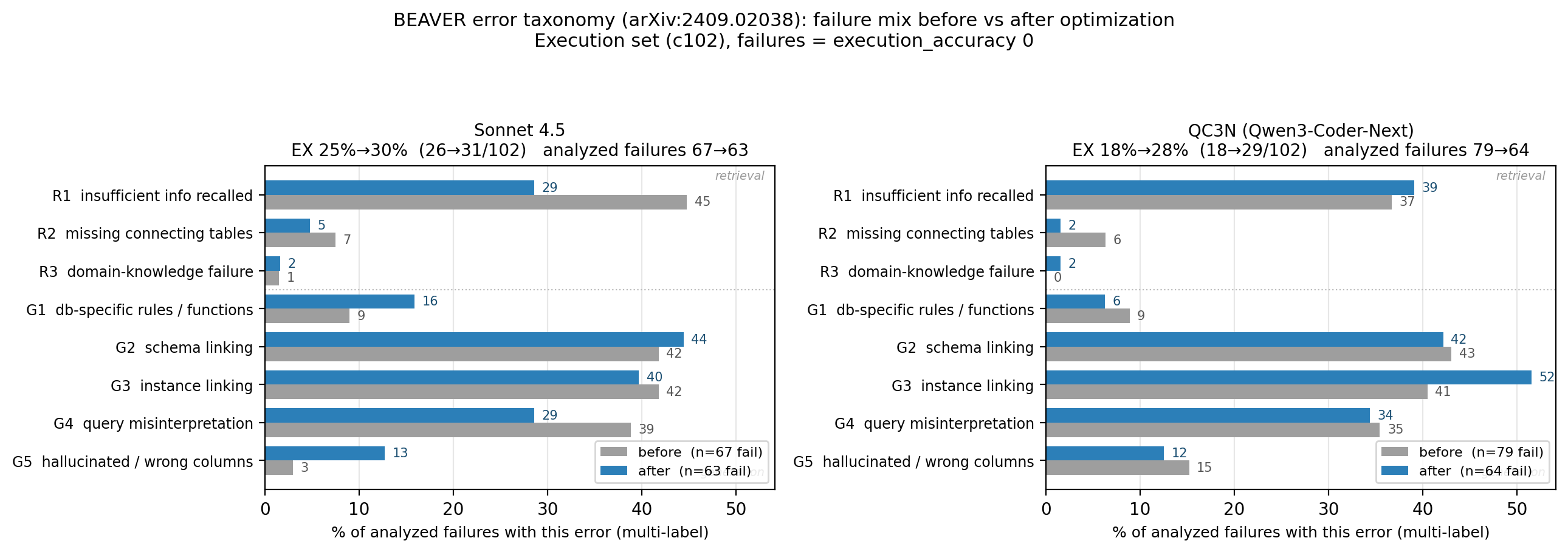}
\caption{BEAVER-taxonomy failure mix before versus after optimization on the internal
execution cohort ($n{=}102$; failures defined by execution accuracy $=0$). Bars show the
percentage of analyzed failures exhibiting each error category (multi-label; bars do not
sum to $100\%$). Retrieval-phase categories (R1--R3) appear above the dotted line and
generation-phase categories (G1--G5) below. Panel titles report binary execution accuracy
and the analyzed-failure count for each condition.}
\label{fig:beaver-taxonomy-errors}
\end{figure}

We analyze residual errors in two complementary ways: by applying the BEAVER error
taxonomy to the executable internal cohort (\Cref{fig:beaver-taxonomy-errors}), and by
query compositionality on BEAVER (\Cref{tab:fail-beaver}). Together, these views agree with
the oracle-graph ceiling of \Cref{sec:diagnosis} and the per-surface attributions the
failure diagnoser (\Cref{sec:error-feedback}) produced during optimization: optimized
cards improve which tables the model reaches for, but the residual difficulty lives in
column-, join-, and composition-level structure.

\paragraph{Internal benchmark: structural error categories.}
Classifying each failing prediction in the executable internal cohort
(\Cref{fig:beaver-taxonomy-errors}), the gains from optimized
cards concentrate in table selection:
\texttt{wrong\_tables} and \texttt{partial\_tables} failures drop and shift into
\texttt{near\_correct}. The residual errors move \emph{downstream} into joins, filters,
and aggregations, the column- and join-level detail the oracle-graph ceiling predicted
would remain difficult.

\textbf{Reliability.} A separate judge on an independent one-in-four sample agreed on the
binary correct/wrong label $89\%$ of the time. Fine-grained codes are noisier, especially
around the R1/G2/G4 boundary, so we treat them as indicative. Cases judged semantically
correct despite EX${=}0$ are excluded, making the analyzed-failure counts slightly smaller
than $102-\mathrm{EX}$.

\textbf{Findings.} Optimization acts primarily on retrieval-phase errors. For Sonnet, R1
insufficient information recalled is the largest single reduction ($45\%\!\to\!29\%$ of
failures, while EX rises from $25\%$ to $30\%$), consistent with optimized summaries
surfacing correct table identifiers so the model stops omitting required tables. By
contrast, generation-phase precision errors persist: schema linking (G2) remains the modal
error in every condition (${\sim}42$--$44\%$), and instance linking (G3) is sticky, rising
as a share of QC3N's smaller residual failure set ($41\%\!\to\!52\%$). In short, the
first-order benefit is better routing to the right evidence: some ``wrong tables'' failures
become ``right tables, wrong columns/predicates'' failures. This independently reproduces
BEAVER's central observation that schema linking is a persistent error class, and it mirrors
our broader result that optimization improves schema grounding before it solves full query
semantics.

\paragraph{BEAVER: difficulty by query compositionality.}
On BEAVER, accuracy is dominated by query structure rather than by which content source is
injected (\Cref{tab:fail-beaver}): single-level (\texttt{base}) queries reach $48\%$, but
deeply compositional (nested-CTE) queries collapse to ${\sim}1.5\%$, uniformly across all
context conditions. Compositionality remains the limiting factor, pointing to
structure-aware generation rather than stronger table-level artifacts as the next lever.

\begin{table}[t]
\centering
\caption{BEAVER execution accuracy by query compositionality ($N{=}300$, \texttt{dw} split).
The pattern holds uniformly across all context conditions of \Cref{tab:beaver}; difficulty is
governed by query structure, not by the injected context.}
\label{tab:fail-beaver}
\footnotesize
\begin{tabular}{lc}
\toprule
Query structure & Exec.\ acc. \\
\midrule
Single-level (\texttt{base})     & $48\%$ \\
Deeply compositional (nested-CTE) & ${\sim}1.5\%$ \\
\bottomrule
\end{tabular}
\end{table}

\section{Example context artifacts}
\label{app:cards}

To make the ablation of \Cref{tab:beaver} concrete, we show the two content sources it
contrasts for one BEAVER table (\texttt{dw.fclt\_rooms}). Both are mined only from
\emph{other} questions' gold SQL under the leakage discipline of \Cref{sec:external};
identifiers and literal values are reproduced verbatim from the source queries.

\paragraph{Aggregated table card (distilled).}
The summarizer $P_{\text{sum}}$ condenses many of the table's historical queries into a single
reusable card with a fixed structure: a one-line purpose, join recipes annotated with
cardinality, observed filter idioms and value domains, a handful of representative verbatim
statements, and explicit correctness rules. Below is the card our optimized $P_{\text{sum}}$
produced for \texttt{dw.fclt\_rooms}, reproduced verbatim and abridged for space (the join-recipe,
two of four example queries, and correctness-rule sections of a six-section card).

\begin{footnotesize}
\begin{verbatim}
# Table SQL Reference Card: dw.fclt_rooms

## 1. Table: dw.FCLT_ROOMS -- Facility room records (room dimensions, access
levels), joined to building tables for aggregating room statistics by building.

## 2. Common Join Recipes + Cardinality
-- FCLT_ROOMS.FCLT_BUILDING_KEY = FCLT_BUILDING.FCLT_BUILDING_KEY
--   (7 examples; 1:many building->rooms -> aggregation required, no DISTINCT when grouping)
-- FCLT_ROOMS.FCLT_ROOM_KEY = COURSE_CATALOG_SUBJECT_OFFERED.MEET_PLACE
--   (1 example; many:many -> requires aggregation by building key)
-- Three-way (3 examples): rooms -> FCLT_BUILDING -> FCLT_BUILDING_ADDRESS
Cardinality notes: all joins fan out building->rooms (1:many); aggregation
(COUNT/AVG/VARIANCE/STDDEV/MIN/MAX) is standard; no DISTINCT under GROUP BY.

## 3. Frequent Filter Idioms + Value Domains
b.BUILDING_TYPE = 'ACADEMIC'   -- (filtered in ALL 12 examples)
b.SITE = 'MIT'                 -- (6 examples)
r.ACCESS_LEVEL IN (1, 2)       -- numeric access level
HAVING COUNT(r.FCLT_ROOM_KEY) > 10 ; HAVING AVG(r.AREA) > 0

## 4. Representative Full-SQL Examples
-- A: statistical aggregation with HAVING
SELECT b.BUILDING_NAME, MAX(r.AREA)-MIN(r.AREA) AS area_range,
       VARIANCE(r.AREA) AS area_variance, STDDEV(r.AREA) AS area_stddev
FROM FCLT_BUILDING b JOIN FCLT_ROOMS r ON b.FCLT_BUILDING_KEY=r.FCLT_BUILDING_KEY
WHERE b.SITE='MIT' AND b.BUILDING_TYPE='ACADEMIC'
GROUP BY b.BUILDING_NAME HAVING COUNT(r.FCLT_ROOM_KEY)>10 ORDER BY area_range DESC;
-- C: three-way join + safe division
SELECT b.BUILDING_NAME_LONG, a.POSTAL_CODE,
       STDDEV(r.AREA)/NULLIF(AVG(r.AREA),0) AS coefficient_of_variation
FROM FCLT_ROOMS r JOIN FCLT_BUILDING b ON r.FCLT_BUILDING_KEY=b.FCLT_BUILDING_KEY
     JOIN FCLT_BUILDING_ADDRESS a ON b.FCLT_BUILDING_KEY=a.FCLT_BUILDING_KEY
WHERE b.BUILDING_TYPE='ACADEMIC' AND b.BUILDING_NAME <> 'ASHDOWN HOUSE'
GROUP BY b.BUILDING_NAME_LONG, a.POSTAL_CODE HAVING AVG(r.AREA)>0
ORDER BY coefficient_of_variation DESC;

## 5. Correctness Rules
- NULL: coefficient of variation MUST use NULLIF(AVG(r.AREA),0) to avoid /0.
- AGG: FCLT_ROOMS is always aggregated when joined (never raw rows); GROUP BY
  on building identifiers; COUNT(r.FCLT_ROOM_KEY) and COUNT(*) interchangeable.
- TYPES: FCLT_BUILDING_KEY compared as string ('32'); ACCESS_LEVEL numeric.
\end{verbatim}
\end{footnotesize}

\paragraph{Raw retrieved query (no distillation).}
The \emph{raw-queries} condition skips the summarizer entirely and injects the retrieved
gold SQL of another \texttt{fclt\_rooms} question verbatim. As \Cref{tab:beaver} shows, this
matches the distilled card to within noise--the model benefits from seeing how the table
is queried, whether or not that signal is first summarized.

\begin{footnotesize}
\begin{verbatim}
WITH inner_cte AS (
  SELECT b.BUILDING_NAME, COUNT(r.FCLT_ROOM_KEY) AS room_count
  FROM FCLT_BUILDING_HIST b JOIN FCLT_ROOMS r ON b.FCLT_BUILDING_KEY = r.FCLT_BUILDING_KEY
  WHERE b.BUILDING_TYPE = 'ACADEMIC' GROUP BY b.BUILDING_NAME )
SELECT BUILDING_NAME, room_count FROM inner_cte
WHERE room_count > ( SELECT AVG(room_count) FROM inner_cte )
ORDER BY room_count DESC LIMIT 10;
\end{verbatim}
\end{footnotesize}

\end{document}